\documentclass[11pt]{article}

\usepackage[preprint]{acl}

\usepackage{times}
\usepackage{xcolor}
\usepackage{booktabs}
\usepackage{latexsym}
\usepackage[listings]{tcolorbox}
\usepackage[T1]{fontenc}

\usepackage[utf8]{inputenc}

\usepackage{todonotes}

\usepackage{microtype}

\usepackage{inconsolata}

\usepackage{graphicx}

\title{Dynamic Role Assignment for Multi-Agent Debate}

\author{
Miao Zhang\textsuperscript{1,2}\thanks{Work done during an internship at Amazon.} \\
\texttt{miaozhng@amazon.com}
\And
Junsik Kim\textsuperscript{2}\footnotemark[2] \\
\texttt{jskimcv@amazon.com}
\And
Siyuan Xiang\textsuperscript{2}\thanks{Equal contribution  with no specific order.} \\
\texttt{siyuanx@amazon.com}
\AND
Jian Gao\textsuperscript{2} \\
\texttt{gajian@amazon.com}
\And
Cheng Cao\textsuperscript{2} \\
\texttt{chengcao@amazon.com}
\AND
\textsuperscript{1} New York University \qquad
\textsuperscript{2} Amazon AGI
}

\begin{document}
\maketitle
\begin{abstract}
Multi-agent large language model (LLM) and vision-language model (VLM) debate systems employ specialized roles for complex problem-solving, yet model specializations are not leveraged to decide which model should fill which role.
We propose dynamic role assignment, a framework that runs a \textbf{Meta-Debate} to select suitable agents before the actual debate. The meta-debate has two stages: (1) proposal, where candidates provide role-tailored arguments, and (2) peer review, where proposals are scored with data and role-specific criteria to choose the best agent for each position. We evaluate our method on LLM problem solving benchmarks. Applied on top of existing debate systems, our approach consistently outperforms uniform assignments (filling all roles with the same model) by up to 74.8\% and random assignments (assigning models to roles without considering their suitability) by up to 29.7\%, depending on the task and the specific assignment. This work establishes a new paradigm for multi-agent system design, shifting from static agent deployment to dynamic and capability-aware selection.
\end{abstract}

\section{Introduction}

\begin{figure}[!t]
\centering
  \includegraphics[width=7.4cm]{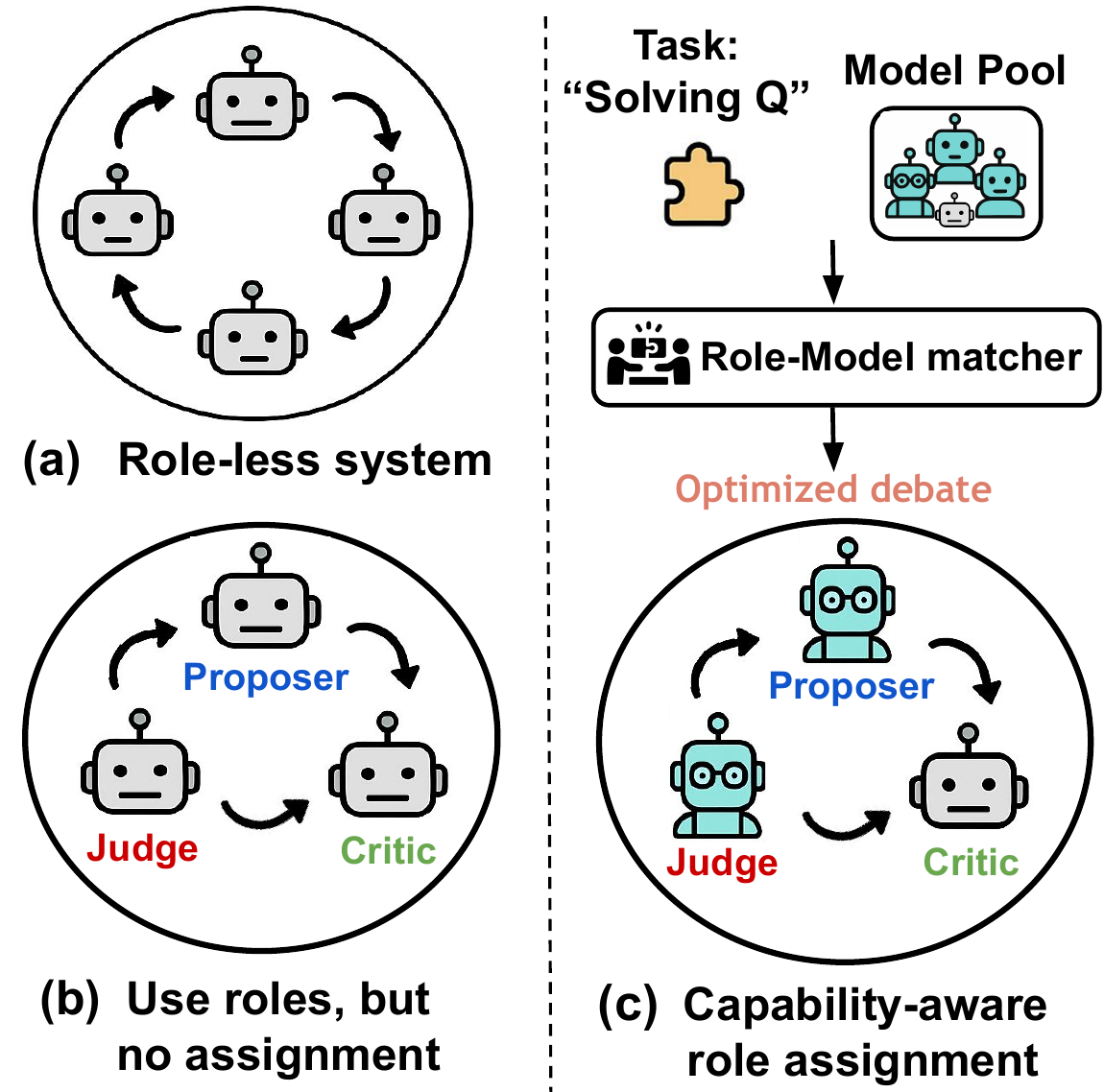}
  \caption{Current LLM based multi-agent debate methods either do not explicitly define agent role functions (a), or use identical model in all roles without assignment (b). In contrast, our method dynamically assigns different models to roles based on tasks (c), matching their strengths to role requirements on each question, leading more effective and optimized collaboration.}
  \label{fig:teaser}
\end{figure}

Multi-agent debate has emerged as a promising approach to enhance the reasoning and decision-making abilities of large language models (LLMs). Instead of relying on a single model, multiple agents generate complementary perspectives and discuss for answers, which helps uncover errors and lead to more reliable final answers (Figure~\ref{fig:teaser} (a)). Recent works have demonstrated that debates can improve accuracy and robustness across a range of complex reasoning tasks, including question answering, translation, dialogue response generation, and LLM-as-a-judge for preference alignment~\cite{du2023improving, chen2024chateval, liu2025breaking, liang-etal-2024-encouraging, li2024improving, zeng2025s2-mad}.

Recent studies highlight the importance of assigning roles to agents in multi-agent debate. When agents have similar capabilities or produce highly similar responses, the debate process often becomes static, increasing risks of reinforcing common misconceptions rather than resolving them~\cite{estornell2024multi}. To address this, recent work has introduced role differentiation strategies, such as assigning agents different collaborative personas~\cite{zhang-etal-2024-exploring, chen2024chateval}, debate functions~\cite{liang-etal-2024-encouraging}, or reasoning methods~\cite{liu2025breaking}, as an example shown in Figure~\ref{fig:teaser} (b). These approaches demonstrate that explicitly modeling role diversity creates complementary reasoning patterns that help identify and correct different types of errors, yielding state-of-the-art performance across LLM reasoning benchmarks.

Despite these advances, existing role-based debate systems typically assume fixed, homogeneous agents, such as using a single state-of-the-art LLM to perform all  roles~\cite{chen2024chateval, liu2025breaking, zhang-etal-2024-exploring}, or rely on role configurations derived from empirical validation results, such as assigning stronger LLMs as debaters and weaker LLMs as the judge~\cite{liang-etal-2024-encouraging, kenton2024scalable}. However, this overlooks the fact that agents may differ in knowledge, reasoning style, or ability for specific roles across different questions. When roles are not matched to agent capabilities, debates can produce adverse effects and erroneously altered original correct answers~\cite{zhang2025if}. Indeed, the impact of role assignment configurations has been shown to vary across individual questions~\cite{zheng2024helpful}. This raises a critical yet underexplored question: given a specific question or task, how can we systematically assign roles to available agents to maximize the overall benefits of debate?



\begin{figure*}[!htbp]
\centering
  \includegraphics[width=\textwidth]{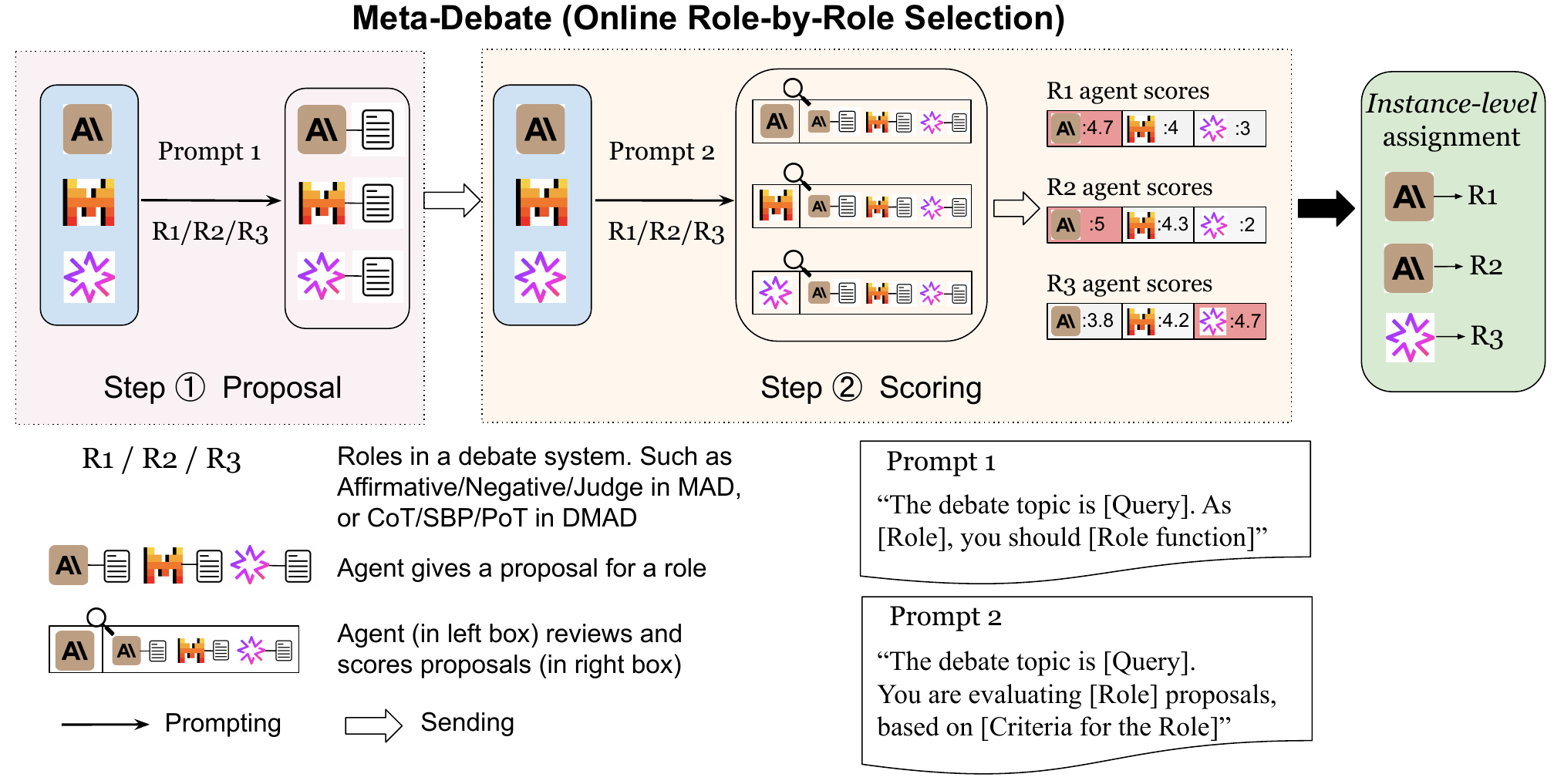}
  \caption{The Meta-Debate framework for role selection consists of two steps. (1) Each agent is first prompted to generate a role-specific response, referred to as a ``proposal''. (2) Then, for each role, all agents evaluate and score the proposals according to predefined criteria. The average of these scores is taken as the final score of each agent for that role for solving the specific question. The agent with the highest score is ultimately assigned to the role, and a same agent can be assigned to multiple roles. }
  \label{fig:pipeline}
\end{figure*}

To address this gap, we propose a dynamic role assignment framework for multi-agent debate (Figure~\ref{fig:teaser} (c)). Specifically, we introduce a Meta-Debate round prior to the main debate, where agents are assigned to roles \textit{adaptively} rather than being fixed in advance. This approach is motivated by the observation that different agents often exhibit varying strengths and knowledge across questions~\cite{luo2023systematic, feng-etal-2024-dont}. In previous debate methods, fixing roles in advance overlooks these strength differences and fails to create a debate team tailored to each problem. In contrast, adaptive role assignment at the question level better matches agents to their strengths, allowing the debate to exploit agent complementary skills and recover correct answers that static setups might miss. Our main contributions are summarized as follows:

\begin{enumerate}
    \item We propose Meta-Debate, a framework for question-level role assignment in multi-agent debate. The method quantifies agent–role suitability to enable tailored role assignment during inference, without relying on labeled data or exhaustive role configuration validations, which is impractical in real-world settings.
    \item We show that debate performance varies substantially across role configurations, highlighting the importance of aligning agents with appropriate roles.
    \item We validate the effectiveness of Meta-Debate on two state-of-the-art debate systems, showing consistent performance gains and offering insights into the design of role-based multi-agent debate.
\end{enumerate}

\section{Related Work}

\paragraph{LLM based multi-agent debate.} 


Multi-agent debates have demonstrated strong performance across reasoning tasks, as they encourage agents to exchange, critique, and refine arguments collaboratively~\cite{du2023improving, chen-etal-2024-reconcile}. It is also common to incorporate roles or personas in debate, which helps fill the “knowledge gap” among agents and diversifies perspectives~\cite{zhang-etal-2024-exploring, feng2024dont}. However, personas do not consistently boost performance, and results depend on role design and agent capabilities~\cite{zheng2024helpful, zhang2025if}. More recently, work has explored deploying mixtures of different LLMs in debates, showing that heterogeneous agents often complement each other with distinct expertise and reasoning styles~\cite{wang2024mixture, tran2025multi}. These findings suggest that assigning roles dynamically to agents with different strengths has great potential, but existing studies rely on static or ad-hoc configurations, leaving the systematic role assignment design largely underexplored.

\paragraph{Automated design of multi-agent systems.} Beyond debate-specific frameworks, several studies have investigated methods to automatically configure multi-agent systems, such as selecting agents based on response confidence~\cite{chen-etal-2024-reconcile}, or by agent influence and contribution to collaboration~\cite{liu2024dynamic} and team diversity~\cite{tian-etal-2025-agentinit}, adaptively routing queries to specialized experts~\cite{wu2025routing}, automating the optimization for instructions and demonstrations in prompt~\cite{opsahl-ong-etal-2024-optimizing}, or designing multi-agent system topology and interaction patterns~\cite{zhou2025multi, zhang2025aflow, wang-etal-2025-beyond, zhang-etal-2025-osc}. These approaches highlight the potential of optimizing agent systems by leveraging diverse model strengths. However, the methods primarily focus on selecting which agents to include in the system or on collaboration mechanisms (e.g., routing and coordination), without addressing the challenge of mapping agents to debate-specific roles. 

Another related line of studies propose role-specific designs. They include reinforcement learning to encourage general debate behaviors among agents (e.g., challenging others, self-reflection)~\cite{li2025advancing}, and role allocators that dynamically generate roles during task execution and assign LLMs to balance performance and cost~\cite{yue-etal-2025-masrouter}. In contrast, our work addresses a complementary but distinct role assignment problem: rather than shaping how agents behave during debate or optimizing performance–cost trade-offs, we study which agents should occupy predefined roles with distinct debate-related reasoning functions, optimizing the alignment between role requirements and agent capabilities.

%


\paragraph{Open challenges.}
To our knowledge, no prior work systematically addresses the problem of automatically matching heterogeneous agents to roles in multi-agent debate. A key difficulty lies in quantifying agent strengths under diverse contexts: an agent may excel in certain domains, modalities, or reasoning styles but underperform in others, making suitability highly context-dependent. Our proposed Meta-Debate framework tackles this gap by incorporating both role requirements and agent capabilities into a dynamic selection process, moving beyond static role assignments toward more adaptive, capability-aware debate configurations. 




\section{Methods}


\subsection{Meta-Debate process}
The overall Meta-Debate framework is illustrated in Figure~\ref{fig:pipeline}. The input consists of a question $Q$, a set of debate roles $R={R_1, R_2, \dots, R_n}$, and a set of available agents $N={N_1, N_2, \dots, N_m}$. The output is a role–agent assignment for the given question, formally represented as a mapping:
\[
\pi_Q : R \;\rightarrow\; N,
\]
where $\pi_Q(R_i) = N_j$ denotes that role $R_i$ is assigned to agent $N_j$ for question $Q$.

In the first step, each agent attempts to fulfill every role and generates a role-specific response, which we refer to as a \emph{proposal}. 
Formally, for each agent $N_j \in N$ and role $R_i \in R$, we denote the proposal as 
\[
P_{i,j}(Q),
\]
representing the response produced by agent $N_j$ when acting in role $R_i$ for question $Q$. All proposals are generated using a uniform prompting template, where the question and role description are inserted (see Prompt~1 in Figure~\ref{fig:pipeline}).
\begin{table*}[t]
  \centering
  \begin{tabular}{lcrrr}
    \toprule
    \textbf{Debate method} & \textbf{Role assignment method} & \textbf{GPQA} & \textbf{MathVision} & \textbf{RealWorldQA} \\
    \midrule
    Single agent (Pixtral)     & NA & 43.08\%  & 32.10\%  & 70.25\%   \\
    Single agent (Claude)     & NA  & 44.42\%  & 30.50\% & 63.61\%     \\
    \midrule
    Multi-agent: MAD   & Pixtral for all roles  & 44.64\% & 31.05\% & 62.24\%     \\
     & Claude for all roles & 54.24\% & 38.42\%  & 63.39\%   \\
    \midrule
     Multi-agent: DMAD    & Pixtral for all roles   & 50.45\% & 28.90\% & 70.94\%      \\
       & Claude for all roles  & 58.93\% & 33.68\% & 67.28\%  \\
    \bottomrule
  \end{tabular}
  \caption{Accuracy results on GPQA, MathVision, and RealWorldQA dataset. We compare single-agent inference and multi-agent debate using the same-model role assignment method (Pixtral or Claude for all roles) across the MAD and DMAD debate frameworks.}
  \label{tab:results_single_homo}
\end{table*}
In the second step, each agent also serves as an \emph{evaluator}, reviewing both their own and other agents' proposals, motivated by the peer review process in human society. 
Formally, for each proposal $P_{i,j}(Q)$ (agent $N_j$ acting as role $R_i$), every agent $N_k \in N$ provides an evaluation score, denoted as
\[
S_{i,j}^{(k)}(Q),
\]
which reflects the assessment of evaluator $N_k$ on the suitability of $N_j$ for role $R_i$. 
The final suitability score of agent $N_j$ for role $R_i$ is then obtained by averaging across all evaluators:
\[
\bar{S}_{i,j}(Q) = \frac{1}{|N|} \sum_{k=1}^{|N|} S_{i,j}^{(k)}(Q).
\]
Finally, each role $R_i$ is assigned to the agent with the highest averaged suitability score:
\[
\pi_Q(R_i) = \arg\max_{N_j \in N} \bar{S}_{i,j}(Q).
\]

Examples of the prompts used and corresponding agent responses for these two steps are provided in the Appendix.

\subsection{Generation of Agent Evaluation Criteria}
The reviewing process is critical to ensure that the agent selected for each role both possesses solid domain knowledge and can effectively fulfill the role’s responsibilities. 
During evaluation, agents refer to the \emph{Criteria for the Role} specified in Prompt 2 (Figure~\ref{fig:pipeline}) to score each proposal. 

To avoid the cost and limitations of manual design, we adopt a data-driven approach to automatically generate role-specific evaluation criteria. 
Specifically, we employ a state-of-the-art LLM
to draft the criteria. 
By providing the data modality and domain (e.g., text-only STEM questions or vision-text scene understanding tasks) along with few illustrative examples, the LLM generates criteria that assess whether a proposal $P_{i,j}(Q)$ from agent $N_j$ demonstrates the knowledge required to solve the target question $Q$. 
Furthermore, by including role-related information defined in prior debate frameworks, the LLM produces criteria that assess whether the agent can effectively fulfill the functional responsibilities of a given role $R_i$. 

In summary, this automated criteria generation module in our Meta-Debate framework tailors agent evaluation to both the problem type and the debate role, leading to more effective assessments than the generic criteria commonly used for LLM response quality.

\begin{table*}[t]
  \centering
  \begin{tabular}{crrr}
    \toprule
    \textbf{Role assignment method} & \textbf{Debate method: MAD} & \textbf{Debate method: DMAD} \\
    \midrule
          Pixtral for all roles  & 44.64\% & 50.45\%     \\
          Claude for all roles   & 54.24\%  & 58.93\% \\
          Nova for all roles    & 52.46\%  & 54.46\%  \\
    \midrule
     Random assignment & 50.67\% & 53.57\%   \\
     Random assignment & 52.23\%  & 58.26\% \\
      Random assignment & 55.58\% & 60.27\%    \\
    \midrule
    Capability-aware assignment (Ours) & \textbf{59.15\%} &  \textbf{66.29\%} \\
    \bottomrule
  \end{tabular}
  \caption{Performance comparison of different role assignment strategies on GPQA dataset. ``Random assignment'' refers to assigning model types (Pixtral/Claude/Nova) to debate roles in MAD or DMAD framework at random.}
  \label{tab:results}
\end{table*}

\section{Experiments}

We evaluate mainstream LLMs as representative agents, including the commercial closed-source model Claude 3.5~\cite{anthropic:claude3.5} (referred to as ``Claude''), Nova Premier~\cite{amazon:nova-premier} (referred to as ``Nova''), and the
open-source model Pixtral-Large-Instruct-2411~\cite{pixtral:large:instruct:2411} (referred to as ``Pixtral''). We use their default settings and hyper-parameters. The temperature is set to 0 to ensure deterministic and replicable outputs across all experiments.

\subsection{Debate baselines}
We consider debate systems with explicit role configurations as suitable baselines for applying our method. 
They include: (1) DMAD~\cite{liang-etal-2024-encouraging} which assigns agents to different reasoning methods as their ``roles.'' 
For text-only problems, the reasoning methods include chain-of-thought (CoT)~\cite{wei2022chain}, step-back prompting (SBP)~\cite{zheng2024take}, and program of thoughts (PoT)~\cite{chen2023program}. 
For vision-text problems, the reasoning methods include input/output standard prompting (IO), compositional chain-of-thought (CCoT)~\cite{mitra2024compositional}, and duty-distinct chain-of-thought (DDCoT)~\cite{zheng2023ddcot}. (2) MAD~\cite{liang-etal-2024-encouraging} which assigns agents to different argument sides, including an Affirmative side to express viewpoints, a Negative side to refute, and a Moderator Judge to draw the final answer. 

In contrast, debate frameworks that do not include role configurations (e.g., SoM~\cite{du2023improving}), or the frameworks that suggest diverse roles or personas but not specify concrete role definitions (e.g., ChatEval~\cite{chen2024chateval}), do not provide the structure needed for our dynamic role assignment method to be applied.

\subsection{Benchmarks}


We evaluate the effectiveness of multi-agent debate systems on three challenging problem-solving benchmarks. These datasets are chosen because they contain questions with verifiable answers, facilitating reliable performance evaluation. Accuracy is used as the metric across all benchmarks.

\textbf{GPQA}~\cite{rein2024gpqa} is a graduate-level multiple-choice benchmark consisting of 448 expert-written questions in biology, chemistry, and physics. We evaluate on the full dataset.

\textbf{MathVision}~\cite{wang2024measuring} is a dataset designed to assess mathematical reasoning in visual contexts, with problems collected from real math competitions. We evaluate all methods and models on the \texttt{testmini} subset and exclude open-ended questions, resulting in 190 evaluation items.

\textbf{RealWorldQA}~\cite{xai2023realworldqa} is a multimodal benchmark designed to assess spatial and contextual understanding in real-world environments. We use the \texttt{test} subset, which contains 437 questions with verifiable answer.

\subsection{Results}

\subsubsection{Model debate competency varies} Table~\ref{tab:results_single_homo} compares single-agent performance with multi-agent debate across the three benchmarks. Overall, multi-agent debate methods generally outperform single-agent settings, consistent with prior findings that debate can leverage diverse knowledge and perspectives to detect errors and enhance reasoning quality~\cite{du2023improving, chen-etal-2024-reconcile, zhang2025critic}. 

However, multi-agent debate does not always guarantee improvement. For example, when all roles in MAD are assigned to Pixtral, performance drops below that of the single-agent Pixtral on both MathVision and RealWorldQA (Table~\ref{tab:results_single_homo}), suggesting that Pixtral struggles to maintain reasoning consistency when taking on specific roles in the MAD framework (Affirmative, Negative, and Moderator Judge). Furthermore, on the RealWorldQA dataset, single-agent Claude performs worse than single-agent Pixtral, but MAD debate using Claude achieves higher accuracy, while DMAD debate using Claude performs worse than that using Pixtral. This indicates that a model’s individual problem-solving ability does not directly translate into its effectiveness in multi-agent debate. Therefore, it is essential to consider each model’s suitability for specific debate roles to ensure effective multi-agent collaboration.

\subsubsection{Role assignment effects}

Table~\ref{tab:results} summarizes the results of baseline role-assignment strategies, including using a single model for all roles and empirical random assignment (randomly selecting a model for each debate role across all samples per benchmark). These baselines are compared against our proposed capability-aware dynamic assignment method, Meta-Debate. Notably, depending on how LLM models are randomly assigned to debate roles, performance can either improve or degrade relative to the vanilla homogeneous setup. The results also exhibit high variance across random configurations, with standard deviation up to 3.5\%. This variation highlights the importance of strategically matching models to roles to fully leverage heterogeneous models and their distinct strengths~\cite{zhang2025if}, while avoiding introducing significant new errors. Overall, random role assignment without guidance is unlikely to yield consistent improvement in practice and may compromise system reliability, underscoring the need for a principled role assignment approach that aligns agent capabilities with role demands and problem requirements.


\subsubsection{Dynamic versus static allocation}
Our proposed Meta-Debate framework addresses these challenges by assigning roles to models based on their quantified suitability for each question, estimated through peer review scores on model proposals. By incorporating both role-specific and question-aware evaluation criteria during peer review, Meta-Debate ensures that each role is filled by the most capable model. As shown in Table~\ref{tab:results}, this adaptive strategy consistently improves accuracy by a substantial margin compared to homogeneous setups (using a single model for all roles) and random role assignments across different debate frameworks. Moreover, Meta-Debate reduces the variance in debate outcomes observed under unguided role assignment, demonstrating enhanced robustness and reliability. These results highlight the value of aligning agent strengths with role demands at the question level and establish dynamic role assignment as a critical step toward more reliable and effective multi-agent debate systems.

\begin{tcolorbox}[colframe=black, colback=blue!2!, coltitle=black, width=\linewidth, boxrule=0.5mm, fontupper=\small]
    \textbf{Question 1:}
    \vspace{0.5em} \\
    \textbf{Evaluator's review on Affirmative proposal 1:}\\
    \texttt{Agent 1 demonstrated an adequate ability to interpret the diagram but \textbf{made an error} in identifying the direction around Buoy 4 and Buoy 5. \textbf{Score is 3}.
    }
    \vspace{0.5em} \\
    \textbf{Evaluator's review on Affirmative proposal 2:}\\
    \texttt{Agent 2 provides precise geometric visualization with a clear explanation of how to determine clockwise motion (keeping buoy on right side). \textbf{Score is 5}.
    }
\end{tcolorbox}

To further demonstrate how Meta-Debate enables more effective debate configurations, we present two representative examples from the MathVision dataset. In each case, Meta-Debate selects the most suitable model for a role based on (1) the model’s knowledge relevant to the question and (2) its ability to fulfill the intended role function. In the first example (Question 1), two models propose to serve as the Affirmative. Proposal 1 contains a reasoning error, whereas Proposal 2 is fully correct. During the Meta-Debate stage, our evaluator assesses reasoning validity and assigns a low score to the flawed proposal (shown in \textbf{bold}), preventing it from being selected. Consequently, the proposal that is both correct and well-reasoned receives a high score and is chosen as the Affirmative, ultimately leading to a correct debate outcome.

\begin{tcolorbox}[colframe=black, colback=blue!2!, coltitle=black, width=\linewidth, boxrule=0.5mm, fontupper=\small]
    \textbf{Question 2:}
    \vspace{0.5em} \\
    \textbf{Negative proposal 1:}\\
    \texttt{As the negative side, \textcolor{red}{I agree with the affirmative side that option D is the correct answer.} Option D correctly shows these circular arcs at the transitions.}
    \vspace{0.5em} \\
    \textbf{Evaluator's review:}\\
    \texttt{Agent 2 simply follows the Affirmative's reasoning and does not provide a different perspective or critique. The agent \textbf{fails to serve as Negative side. Score is 2.}
    }
    \vspace{0.5em} \\
    \textbf{Negative proposal 2:}\\
    \texttt{\textcolor{blue}{I disagree with Affirmative’s conclusion.} The locus of the midpoint of the wheel is not the same as the locus of the center...The answer should be E.}
    \vspace{0.5em} \\
    \textbf{Evaluator's review:}\\
    \texttt{Agent 3 raises valid critique by pointing out Affirmative's error in interpretation the question, therefore suitable for Negative role. \textbf{Score is 5}.
    }
\end{tcolorbox}

In the second case (Question 2), when selecting the Negative role, Proposal 1 submitted by one model fully agrees with the Affirmative’s incorrect reasoning and fails to provide new insights (shown in \textcolor{red}{red}). Meta-Debate identifies this weakness (in \textbf{bold}) and assigns a low score, as the proposal does not fulfill the intended function of the Negative role. In contrast, Proposal 2 generated by a different model presents a critical counter-argument that correctly identifies flaws in the Affirmative’s reasoning and offers the correct perspective (in \textcolor{blue}{blue}). This case illustrates that different LLMs exhibit distinct knowledge bases and abilities to fulfill specific roles, reinforcing the importance of dynamic role assignment in collaborative systems. Moreover, our method benefits from evaluation criteria that consider both content quality (i.e., reasoning validity and correctness) and role effectiveness (i.e., whether the agent fulfills the expected function of its role).

\section{Conclusion}

This work introduces Meta-Debate, a framework that dynamically assigns debate roles based on agents’ demonstrated capabilities. Observing that agents differ in knowledge and reasoning strengths across tasks, Meta-Debate adopts a two-stage process: it first elicits role-specific proposals from candidate agents, then uses peer review with data- and role-aware evaluation criteria to select the optimal configuration. This approach ensures debates are both diverse in perspectives and strategically aligned with participant strengths.

Experiments on challenging benchmarks spanning STEM reasoning and multimodal understanding show that Meta-Debate substantially outperforms single-agent and static multi-agent baselines. By aligning agent abilities with role demands, Meta-Debate improves reasoning robustness across text and vision-language tasks, marking an important step toward more reliable and adaptive multi-agent systems.

\section{Limitations}

While Meta-Debate demonstrates strong improvements, it has two main limitations. First, its effectiveness is bounded by the fundamental capabilities of the underlying agents. If none of the available agents possess the necessary knowledge or reasoning ability to solve a question in any roles, no role assignment strategy can overcome this deficiency. Second, Meta-Debate introduces additional computational and token costs, as it requires extra steps for prompting agents to generate and evaluate proposals for each role. In future work, these costs may be reduced by introducing an adaptive control mechanism that decides when to perform Meta-Debate. For example, the system could bypass Meta-Debate for overly easy questions where any debate configuration suffices, or for overly difficult questions that are unsolvable given current agent limitations. Such an adaptive module would preserve the benefits of dynamic role assignment while improving efficiency.


\bibliography{custom}

\appendix

\section{Appendix}
\label{sec:appendix}

\subsection{Examples of prompts used in Meta-Debate}

1. Prompt 1 in Figure~\ref{fig:pipeline} which lets agents to generate proposals for ``Affirmative'' role in MAD:

\texttt{You are a debater. Hello and welcome to the debate competition. It’s not necessary to fully agree or disagree with each other’s perspectives, as our objective is to find the correct answer. The debate topic is to solve a given question and select the correct answer. Here is the data:}

\texttt{Question:}

\texttt{Choices:}

\texttt{You are affirmative side. Please express your viewpoints.}


1. Prompt 2 in Figure~\ref{fig:pipeline} which lets agents to evaluate and generate scores for the three candidates for ``Affirmative'' role in MAD:

\texttt{You are participating in a debate assessment to evaluate agent performance on a given STEM question. Here is the data:}

\texttt{Question:}

\texttt{Choices:}

\texttt{Context: This is part of a structured debate with three roles (Affirmative, Negative, and Moderator Judge). }

\texttt{Three agents have each submitted responses attempting to fulfill the Affirmative role.}

\texttt{Agent 1 Response:}

\texttt{Agent 2 Response:}

\texttt{Agent 3 Response:}

\texttt{Your Task: Evaluate each agent's suitability for the Affirmative role based on their submitted responses using the criteria below. You will provide a 1–5 score for each criterion per agent, along with a brief explanation.}

\texttt{Evaluation Criteria:}

\texttt{1. Accuracy (1-5) - The response demonstrates correct application of fundamental principles, proper use of formulas/equations, and arrives at the right answer through valid reasoning steps without mathematical or conceptual errors.}

\texttt{2. Technical Depth (1-5) - The response shows genuine understanding of underlying mechanisms and theory, providing substantive analysis that goes beyond surface-level observations to address the core scientific concepts involved.}

\texttt{*Note*: Do not be biased toward agent writing styles (e.g., length, complexity)}

\texttt{Scoring Scale:}

\texttt{5 = Excellent - Natural fit, strong performance on this criterion}

\texttt{4 = Good - Minor room for improvement}

\texttt{3 = Adequate - Acceptable but with clear concerns}

\texttt{2 = Poor - Significant weaknesses}

\texttt{1 = Very Poor - Clearly inadequate performance}

\subsection{Prompt examples for criteria generation}
\textbf{Context:}
The debate system includes an Affirmative debater to first express viewpoints, a Negative debater to respond to Affirmative's answer and correct the mistakes. A Judge to review Affimrative and Negative's arguments, then decides whether the correct solution can be obtained and choose a side to support.
Several agent candidates have played the Affirmative role to answer visual math questions. Below are example questions:

Question1:
\{question1\}

Question2:
\{question2\}

Question3:
\{question3\}

\textbf{Your Task:}
Draft 2 or 3 evaluation criteria that can be used to judge which agent gives the best response being Affirmative. Each criterion should be a single word or short phrase, followed by a brief explanation.

Your criteria should focus on:
\begin{itemize}
    \item If the agent fullfills the Affirmative role
    \item Correctness and validity of the agent reasoning rather than presentation styles
\end{itemize}

The criteria should reflect what distinguishes high-quality, technically correct Affirmative reasoning from shallow and flawed attempts. The criteria should only include necessary traits for the goal of answering correctly for visual math problems.
The criteria should not mention the given example questions. You should only output the criteria.

\end{document}